# Reinforcement Learning by Comparing Immediate Reward

Punit Pandey
Department Of Computer Science and Engineering,
Jaypee University Of Engineering And Technology

DeepshikhaPandey
Department Of Computer Science and Engineering,
Jaypee University Of Engineering And Technology

Dr. Shishir Kumar
Department Of Computer Science and Engineering,
Jaypee University Of Engineering And Technology

*Abstract*— **This paper introduces an approach to Reinforcement Learning Algorithm by comparing their immediate rewards using a variation of Q-Learning algorithm. Unlike the conventional Q-Learning, the proposed algorithm compares current reward with immediate reward of past move and work accordingly. Relative reward based Q-learning is an approach towards interactive learning. Q-Learning is a model free reinforcement learning method that used to learn the agents. It is observed that under normal circumstances algorithm take more episodes to reach optimal Q-value due to its normal reward or sometime negative reward. In this new form of algorithm agents select only those actions which have a higher immediate reward signal in comparison to previous one. The contribution of this article is the presentation of new Q-Learning Algorithm in order to maximize the performance of algorithm and reduce the number of episode required to reach optimal Q-value. Effectiveness of proposed algorithm is simulated in a 20 x20 Grid world deterministic environment and the result for the two forms of Q-Learning Algorithms is given.**

## I. INTRODUCTION

Q-Learning algorithm proposed by Watkins [2,4] is a model free and online reinforcement learning algorithm. In reinforcement learning selection of an action is based on the value of its state using some form of updating rule. There is an interaction between agent and environment where the agent has to go through numerous trials in order to find out the best action. An agent chooses that action which has maximum reward obtained from its environment. The reward signal may be positive or negative depends on the environment.

Q-learning has been used in many applications because it does not require the model of environment and is easy to implement. State-action value, a value for each action from each state, converges to the optimal value as state-action pairs are visited many times by the agent.

In this article we propose a new relative reward strategy for agent learning. Two different form of Q-Learning method is considered here as a part of study. First form of Q-Learning method uses a normal reward signal. In this algorithm Q-value evaluates whether things have gotten better or worse than expected as a result of an action selection in the previous state. The action selected by agents is most favorable which has lower TD error. Temporal difference is computed on the basis of normal reward gain by agents from its surroundings. An estimated Q-value in the current state is than determined using Temporal Difference. Agent actions are generated using the maximum Q-values. The second form of Q-Learning algorithm is an extension towards a relative reward. This form of Q-Learning method utilizes the relative reward approach to improve the learning capability of algorithm and decreases the number of iteration. In this algorithm only those action is selected which has a better reward from its previous one.

This idea comes from psychological point of views that human beings tend to select only those action which has higher reward value. However, this algorithm is not suitable for multi agent problems. To demonstrate effectiveness of the proposed Q-Learning algorithm, java applet is utilized to simulate a robot that reaches to a fixed goal. Simulation result confirms that the performance of proposed algorithm is convincingly better than conventional Q-learning.

This paper is organized as follows: Basic concept of reinforcement learning is presented in section 2. Section 3 describes about the conventional Q-Learning method. Section 4 presents a new Relative Q-Learning in context of relative immediate reward. Section 5 describes Experimental setup & results and concluding remarks follow in Section 6.

## II. REINFORCEMENT LEARNING

Reinforcement learning (RL) is a goal directed learning methodology that is used to learn the agents. In Reinforcement learning [1,5,6,7,8,9] the algorithm decide what to do and how to map situations to actions so that we maximize a numerical reward signal. The learner is not advised which actions to take, but instead it discover which actions provide the maximum reward signal by trying them. Reinforcement learning is defined by characterizing a learning problem. Any algorithm that can able to solve the defined problem, we consider to be a reinforcement learning algorithm. The key feature of reinforcement learning is that it explicitly considers the whole problem of a goal-directed agent interacting with an uncertain environment. All reinforcement learning agents [3,10,11,12] have explicit goals, can sense aspects of their environments, and can choose actions to influence their environments. In reinforcement learning agent prefer to choose actions that it has tried in the past and found to be effective in producing maximum reward. The agent has to exploit based on what it already knows in order to obtain reward and at the same time it also has to explore in order to make better action selections in the future. Reinforcement learning has four elements policy, reward function, value function and model of environment.

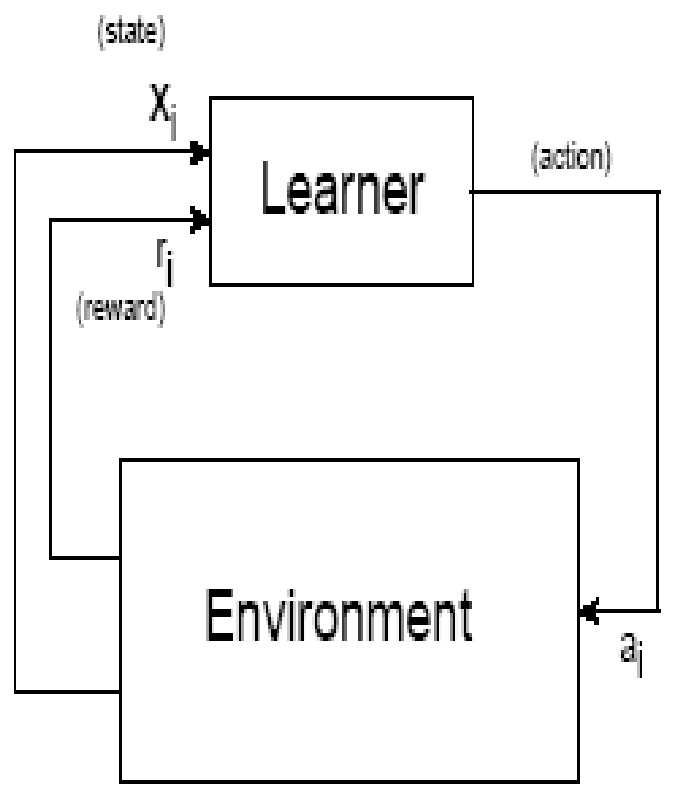

Figure 1.1:.Reinforcement learning

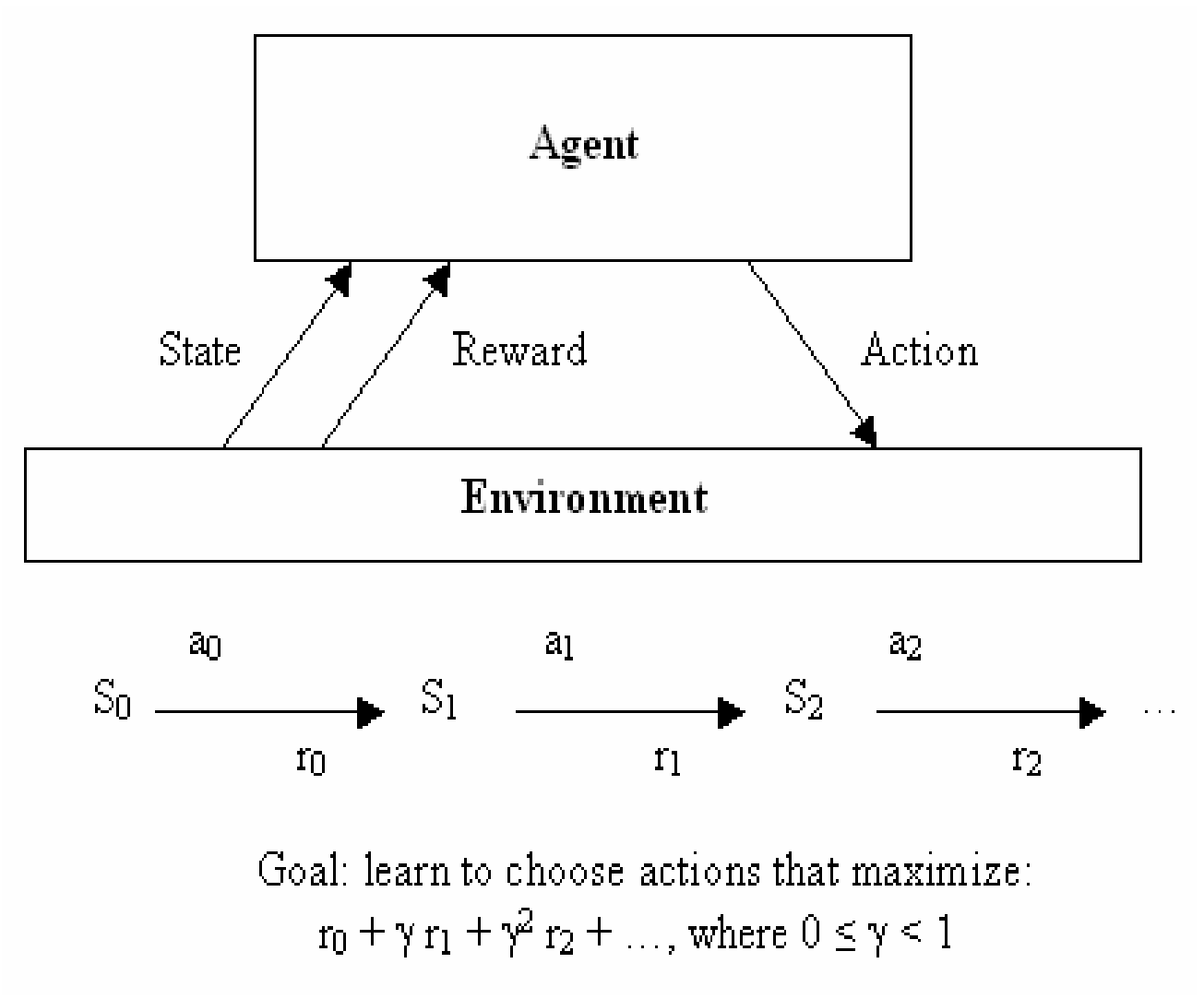

Figure 1.2:Reinforcement learning

Model of the environment is an optional element because reinforcement learning also supports the model free algorithms like Q-learning.

A policy for our agent is a specification of what action to take for every Input. In some cases policy may be a simple function or look-up table or sometime it can be an extensive computation. The policy is the core of reinforcement learning agent because it alone is sufficient to take the decision on further action.

## III. Q-LEARNING

Q-learning is a form of model-free reinforcement learning [2] (i.e. agent does not need an internal model of environment to work with it). Since Q-learning is an active reinforcement technique, it generates and improves the agent's policy on the fly. The Q-learning algorithm works by estimating the values of state-action pairs.

The purpose of Q-learning is to generate the Q-table, Q(s,a), which uses state-action pairs to index a Q-value, or expected utility of that pair. The Q-value is defined as the expected discounted future reward of taking action a in state s, assuming the agent continues to follow the optimal policy. For every possible state, every possible action is assigned a value which is a function of both the immediate reward for taking that action and the expected reward in the future based on the new state that is the result of taking that action. This is expressed by the one-step Q-update equation [2,4,10,13,14].

$$Q(s, a) \leftarrow Q(s, a) + \alpha\, [r + \gamma * \max_{a'} Q(s', a') - Q(s, a)] \quad (1)$$

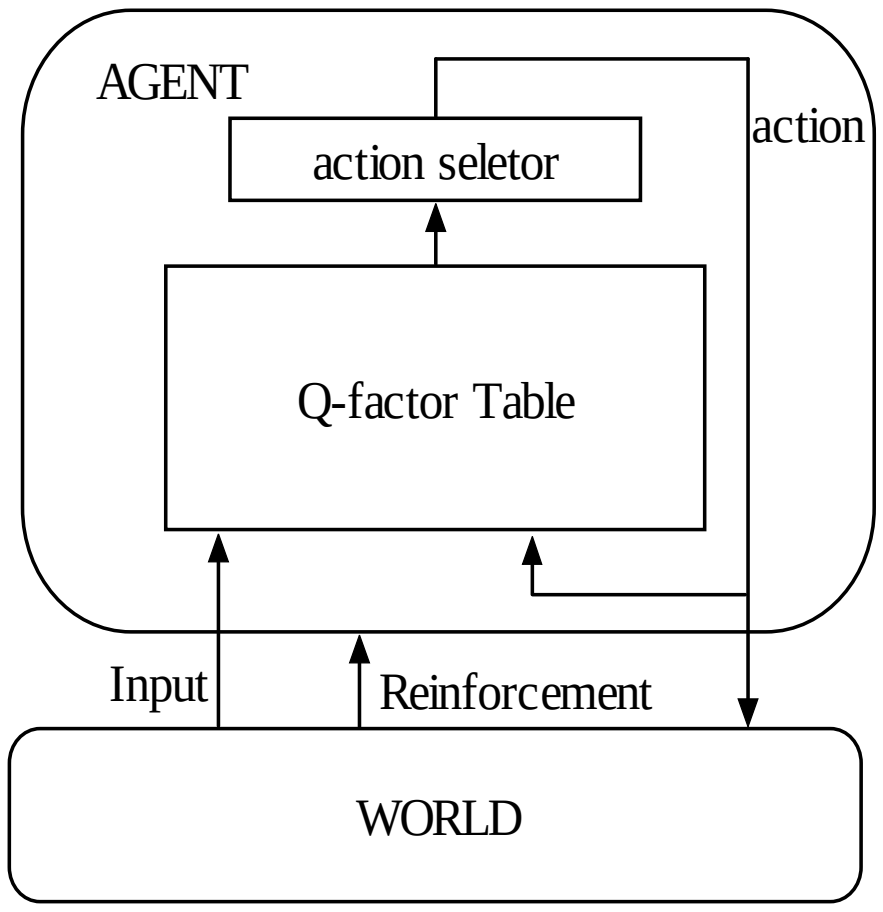

Figure 2: Structure of the Q-Learning agent

Where α is the learning factor and γ is the discount factor. These values are positive decimals less than 1 and are set through experimentation to affect the rate at which the agent attempts to learn the environment. The variables s and a represent the current state and action of the agent, r is the reward from performing s' and a', the previous state and action, respectively.

The discount factor makes rewards earned earlier more valuable than those received later. This method learns the values of all actions, rather than just finding the optimal policy. This knowledge is expensive in terms of the amount of information that has to be stored, but it does bring benefits. Q-learning is exploration insensitive, any action can be carried out at any time and information is gained from this experience. The agent receives reinforcement or reward from the world, and returns an action to the world round and round as shown below:

### A. *Elementary parts of Q-learning:*

*Environment:*

Q-learning based on model-free mode of behavior i.e the environment is continuously changing. Agent does need to predict future state. Environment can be either deterministic or non-deterministic. In deterministic environment application of single state lead to a single state where as in nondeterministic environment application of a single action may lead to a number of possible successor states. In case of non-deterministic environment, each action not only labeled with expected immediate reward but also with the probability of performing that action. For the sake of simplicity we are considering deterministic environment in this thesis work.

*Reward Function:*

A reward function defines the goal in a reinforcement learning problem. it maps each perceived state (or state-action pair) of the environment to a single number, a reward, indicating the intrinsic desirability of that state. A reinforcement learning agent's sole objective is to maximize the total reward it receives in the long run. The reward function defines what the good and bad events are for the agent.

*Action-value function:*

The Q-learning learning is based upon Quality-values (Q-values) Q(s,a) for each pair (s,a). The agent must cease interacting with the world while it runs through this loop until a satisfactory policy is found. Fortunately, we can still learn from this. In Q-learning we cannot update directly from the transition probabilities-we can only update from individual experiences. In 1 step Q-learning, after each experience, we observe state s', receive reward r, and update:

$$Q(s, a) = r + \gamma \max_{a'} Q(s', a') \quad (2)$$

### B. *Q-learning Algorithm:*

Initialize Q(s, a) arbitrarily

Repeat (for each episode)

Choose a starting state, s

Repeat (for each step of episode):

Choose a from s using policy derived from Q

Take action a, observe a immediate reward r, next state s'

$Q(s, a) \leftarrow Q(s, a) + \alpha [r + \gamma * \max_{a'} Q(s', a') - Q(s, a)]$

$s \leftarrow s'$ ;

Until state s' match with the Goal State

Until a desired number of episodes terminated

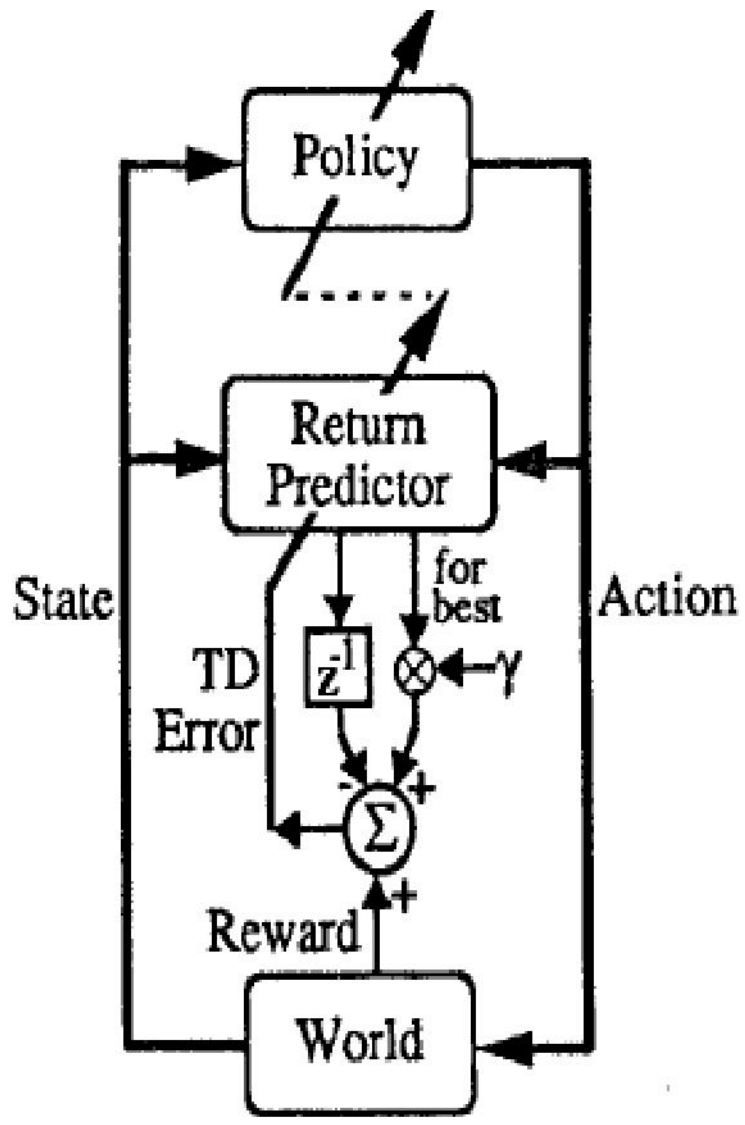

Figure 3: Q-Learning Architecture

## IV. RELATIVE Q-LEARNING

This section introduces a new approach Relative reward to conventional Q-learning that makes Relative Q-Learning. Conventional Q-learning has been shown to converge to the optimal policy if the environment is sampled infinitely by performing a set of actions in the states of the environment under a set of constraints on the learning rate α. No bounds have been proven on the time of convergence of the Q-learning algorithm and the selection of the next action is done randomly when performing the update. This simply mean that the algorithm would take a longer time to converge as a random set of states are observed which may or may not bring the state closer to the goal state. Furthermore, it means that this function cannot be used for actually performing the actions until it has converged as it has a high chance of not having the right value as it may not have explored the correct states. This is especially a problem for environments with larger state spaces. It is difficult to explore the entire space in a random fashion in a computationally feasible manner. So by applying below mention method and algorithm we try to keep the Q-learning algorithm near to its goal in less time and less number of Episode.

### A. *Relative Reward*

Relative reward is a concept that compares (current reward with the previous received reward) two immediate rewards. The objective of the learner is to choose actions maximizing discounted cumulative rewards over time. Let there is an agent in state st at time t, and assume that he chooses action at. The immediate result is a reward rt received by the agent and the state changes to st+1. The total discounted reward [2,4] received by the agent starting at time t is given by:

$$r(t) = r_t + \gamma r_{t+1} + \gamma^2 r_{t+2} + \dots\dots\dots + \gamma^n r_{t+n} + \dots\dots\dots \quad (3)$$

Where γ is discount factor in the range of (0:1).

The immediate reward is based upon the action or move taken by an agent to reach the defined goal in each episode. The total discounted reward can maximize in less number of episode if we select the higher immediate reward signal from previous.

### B. Relative Reward based Q-Learning Algorithm

Relative reward based Q-learning is an approach towards maximizing the total discounted rewards. In this form of Q-learning we selected the maximum immediate reward signal by comparing it with previous one. This is expressed by the new Q-update equation.

$Q(s, a) = Q(s, a) + \alpha\, [\max(r(s,a),r(s',a')) + \gamma\ \max_{a'} Q(s', a') - Q(s, a)]$

Algorithm:

Initialize Q(s, a) arbitrarily

Repeat (for each episode)

Choose a starting state, s

Repeat (for each step of episode):

Choose a from s using policy derived from Q

Take action a, observe a immediate reward r, and next state s'

$Q(s, a) = Q(s, a) + \alpha\, [\max(r(s,a),r(s',a')) + \gamma\ \max_{a'} Q(s', a') - Q(s, a)]$

$s \leftarrow s'$ ;

Until state s' match with the Goal State

Until a desired number of episodes terminated

## V. Experiments & Results

The Proposed Relative Q-Learning was tested on 10 x 10 and 20 x 20 grid world environment. In the Grid World Square There are four possible actions for the agent as it is a deterministic environment given in figure 4.

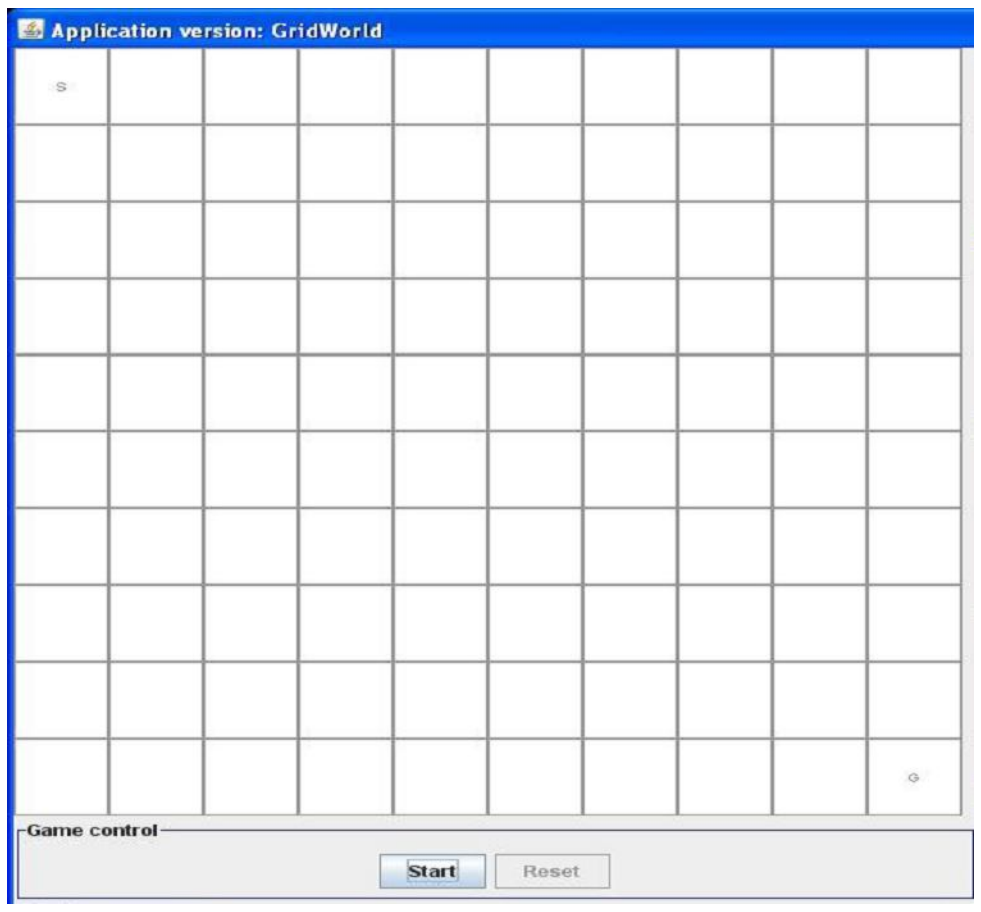

Figure4: A 10 x 10 Grid World Environment

In order to consider the situation of encountering a wall, the agent has no possibility of moving all the way in the given direction. When the agent enters into goal states, it receives 50 as a reward. We are also providing the immediate reward value by incrementing or decrementing the Q-value marked with S represent the start state and G represent the goal state. The purpose of the agent is to find out the optimum path to arrive at the goal state starting from the start state, and to maximize the reward it receives.

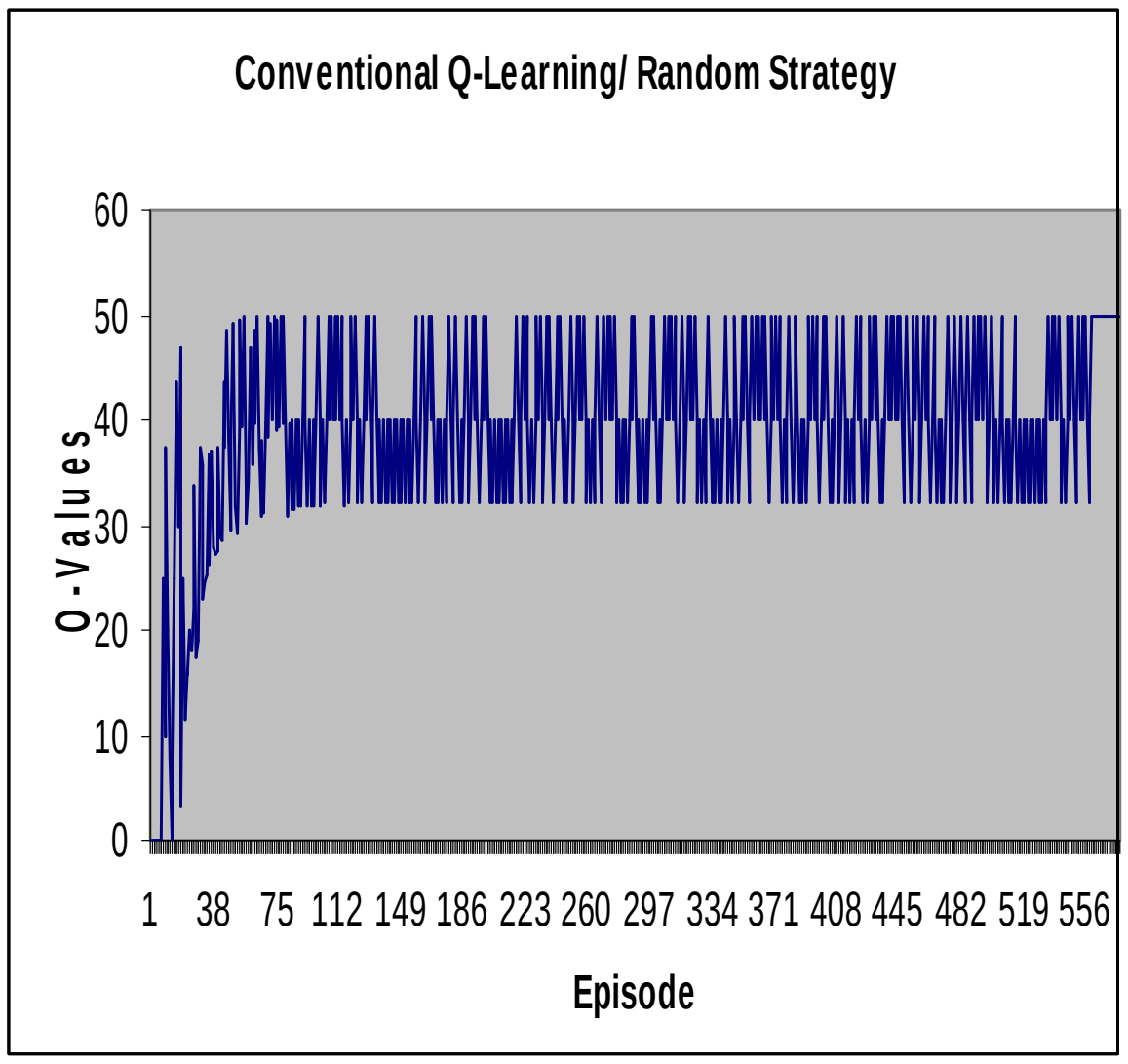

Figure5: Conventional Q-Learning.

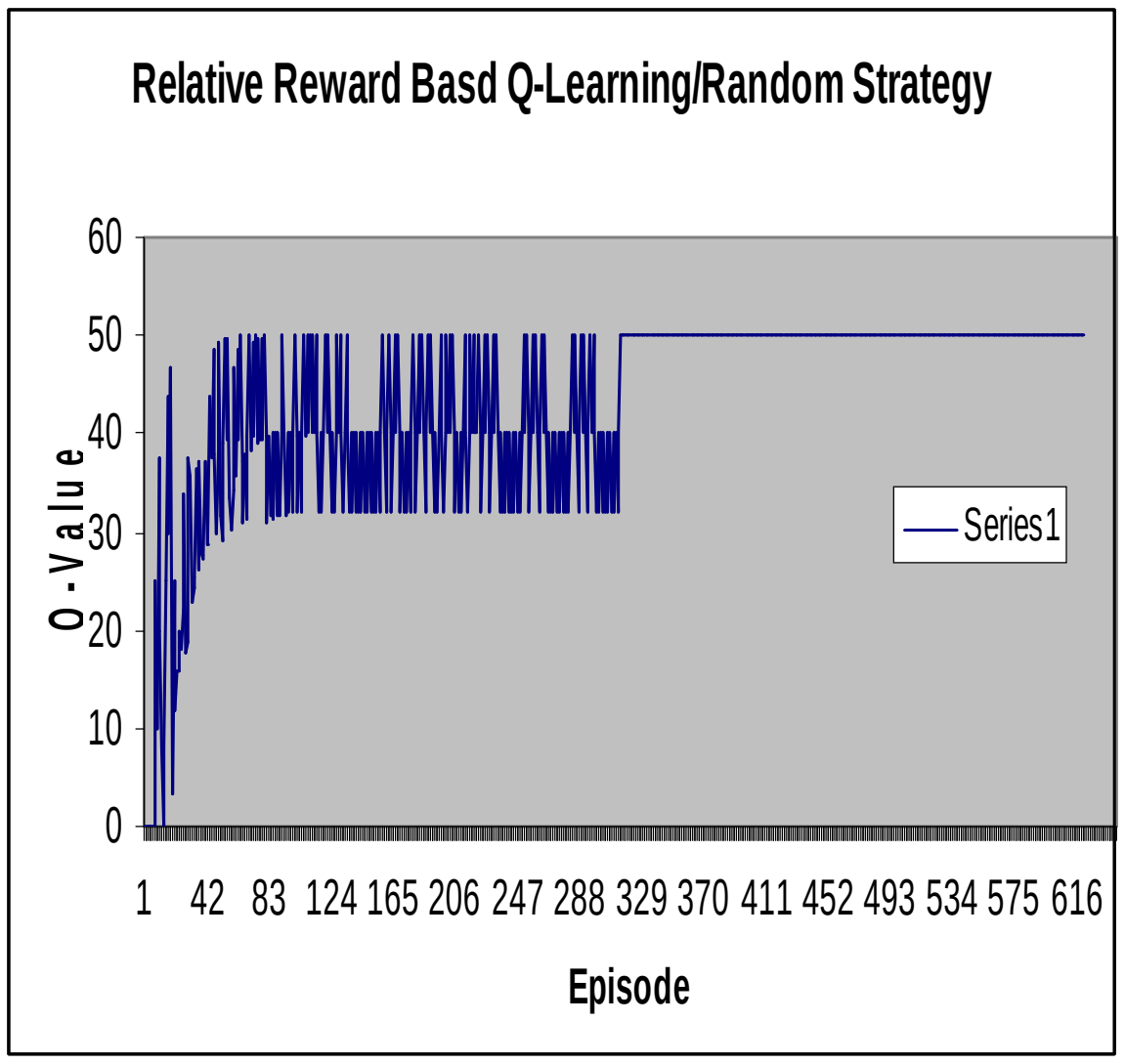

Figure6: Relative Q-Learning

We have executed 500 episodes to converge the Q-value. The grid world is a deterministic environment so the value of learning α and discount rate Y were set to 0.8. Figure 5 &

Figure 6 shows the relationship between Q-Values and the number of episode where x axis represents the number of episode and y axis represents the Q-values. Figure 5 represents the result of conventional Q-Learning where we can see that Q-value converges after executing 500 episodes but in figure 6 Relative Q-learning takes 300 episode

So we can say that convergence rate of relative Q-learning is faster than conventional Q-learning.

## VI. CONCLUSION & FUTURE WORK

This paper proposed an algorithm which compares the immediate reward signal with its previous one. The agent will immediately return back to previous state if it will receive the lower reward signal for that particular move. If conventional Q-learning was applied in the real experiment, a lot of iterations were required to reach the optimal Q values. The Relative Q-learning algorithm was proposed for environment which used small amount of episodes to reach the convergence of Q-values. This new concept allows the agent to learn uniformly and helps in such a way so that it will not deviate from its goal. Part of future work may be included to verify the proposed algorithm in nondeterministic environment.